\documentclass{article}
\usepackage{amsmath,epsfig}
\usepackage[preprint]{spconfa4}
\usepackage{xcolor}
\usepackage{cite}

\usepackage{graphicx}
\usepackage{booktabs}
\usepackage{multirow}

\let\OLDthebibliography\thebibliography
\renewcommand\thebibliography[1]{
  \OLDthebibliography{#1}
  \setlength{\parskip}{0pt}
  \setlength{\itemsep}{0pt plus 0.3ex}
}

\newcommand{\etal}{{\emph{et al.}}}

\begin{document}\sloppy

\def\x{{\mathbf x}}
\def\L{{\cal L}}

\title{PanFormer: a Transformer Based Model for Pan-sharpening}
%
\name{Huanyu Zhou$^1$, Qingjie Liu$^{1,2,}$\sthanks{Corresponding author} and Yunhong Wang$^1$}
\address{$^1$The State Key Laboratory of Virtual Reality Technology and Systems, Beihang University, Beijing China\\ $^2$Hangzhou Innovation Institute, Beihang University, Hangzhou, China\\ \{zhysora, qingjie.liu, yhwang\}@buaa.edu.cn}

\maketitle

\begin{abstract}
Pan-sharpening aims at producing a high-resolution (HR) multi-spectral (MS) image from a low-resolution (LR) multi-spectral (MS) image and its corresponding panchromatic (PAN) image acquired by a same satellite. Inspired by a new fashion in recent deep learning community,  we propose a novel Transformer based model for pan-sharpening. We explore the potential of Transformer in image feature extraction and fusion. Following the successful development of vision transformers, we design a two-stream network with the self-attention to extract the modality-specific features from the PAN and MS modalities and apply a cross-attention module to merge the spectral and spatial features. The pan-sharpened image is produced from the enhanced fused features. Extensive experiments on GaoFen-2 and WorldView-3 images demonstrate that our Transformer based model achieves impressive results and outperforms many existing CNN based methods, which shows the great potential of introducing Transformer to the pan-sharpening task. Codes are available at https://github.com/zhysora/PanFormer.
\end{abstract}
\begin{keywords}
Pan-sharpening, transformer, attention mechanism, remote sensing
\end{keywords}
\section{Introduction}
\label{sec:intro}

High-quality remote sensing images with both high spatial and spectral resolutions are demanded in many practical applications, such as geography, land surveying, and environment monitoring. Most remote sensing sensors provide images in a pair of modalities: the multi-spectral (MS) images and their corresponding panchromatic (PAN) images. MS images are with rich spectral information however a lower spatial resolution, while PAN images have high spatial resolution however with only one band. To combine the strengths of both modalities, the pan-sharpening task focuses on merging the complementary information from PAN and MS images and creating the high-resolution MS (HR MS) images.

Recently, deep learning has achieved great successes in various computer vision tasks \cite{ICME_SR1, ICME_SR2}, motivating researchers to develop pan-sharpening methods that leverage the power of deep neural networks. PNN \cite{PNN}, who borrows the idea from a super-resolution network \cite{SRCNN}, designs a 3-layer CNN to solve the pan-sharpening. This work inspired a number of subsequent studies \cite{PanNet, DRPNN, MSDCNN}. However, most of them address pan-sharpening from an image super-resolution view that learns non-linear mappings from the joint PAN-MS space to the target HR MS space, in which the PAN image is regarded as a channel of the input. Considering the distinctions between PAN and MS images carry distinct information, some researchers propose to extract features of the two modalities using different subnetworks \cite{Pan-GAN, DiCNN, TFNet}. 

It has been found that capturing the correlations between the PAN and the MS bands and incorporating them into the fusion process is essential for reducing spectral distortions of HR MS \cite{BDSD, datafitting}. However, few works have taken this into account. In this paper, we address this problem with a novel Transformer \cite{transformer} based network. Our method introduces a novel cross-attention block that is capable of modeling redundant and complementary information across the PAN and MS modalities. We first construct an encoder with a transformer architecture to extract modality-specific features from PAN and MS images, respectively. Then, we propose a cross-attention operation to encourage the information exchange between the two modalities. The cross-attention module can capture complex correlation relationships between the PAN and MS, which is crucial for achieving an optimal fusion performance. In the head, we apply a restoration module to generate the final pan-sharpened images. 

Our contributions are summarized as follows: 1) We design a Transformer based model for pan-sharpening, termed PanFormer. To the best of our knowledge, it is the first time that Transformer is applied to the pan-sharpening task. 2) We design a two-stream network to extract modality-specific features and fuse them with a novel cross-attention module. The cross-attention module is capable of capturing redundant and complementary information of the PAN and MS modalities, thus enabling a good pan-sharpening performance. 3) Experiments on GaoFen-2 and WorldView-3 datasets demonstrate that our proposed method presents competitive performance compared to existing CNN based models. 

\begin{figure*}[htb]
  \includegraphics[width=\linewidth]{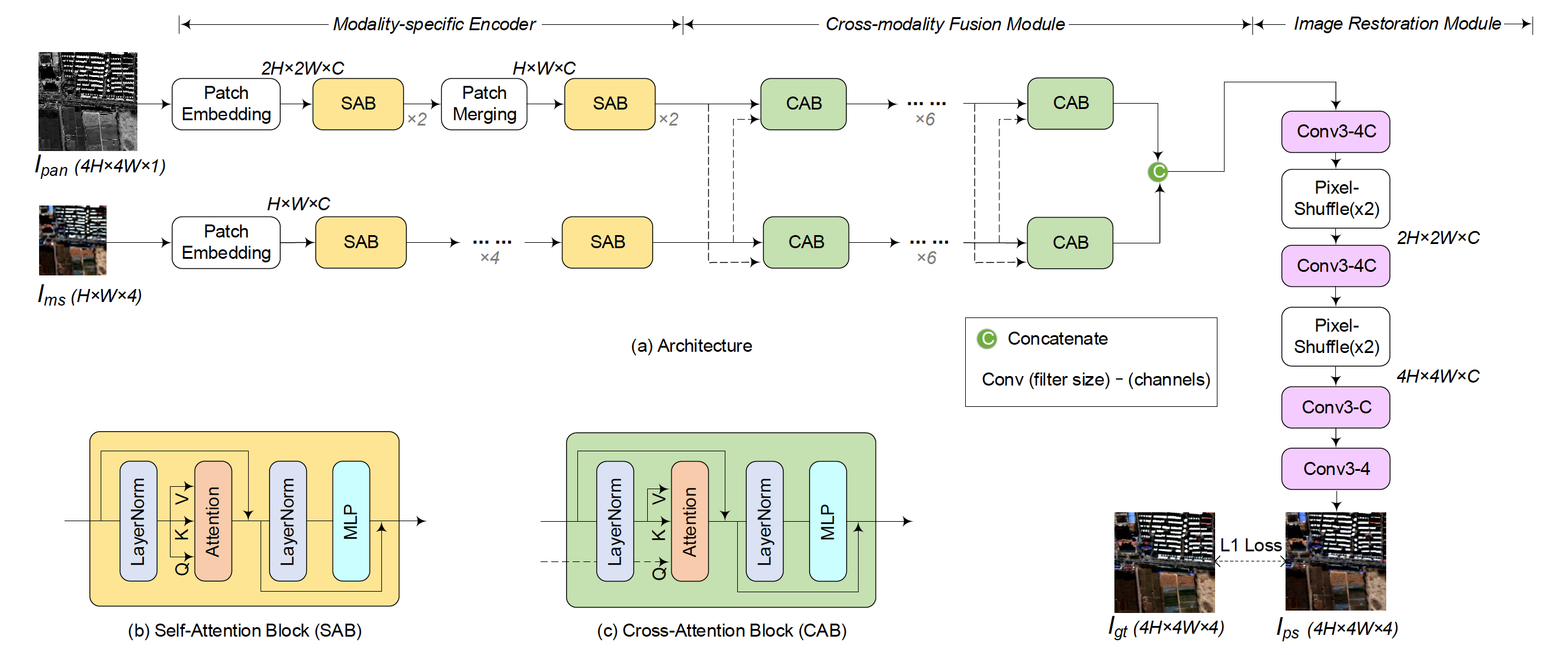} 
  \caption{The architecture of PanFormer. PanFormer consists of 3 modules: the modality-specific encoder, the cross-modality fusion module, and the image restoration module. It is noted that PanFormer is trained only with L1 loss.}
  \label{fig::architecture}
  \vspace{-10pt}
\end{figure*}


\section{Related Work}

\subsection{Deep learning based pan-sharpening}
Recently, deep learning techniques have entered a booming period and shown dominating performances in various computer vision fields. Observing that the pan-sharpening task shares a similar spirit with super-resolution, Masi \etal \cite{PNN} borrow the idea from SRCNN \cite{SRCNN} and propose a 3-layer CNN model to solve the pan-sharpening. To further improve the performance, more and more researchers devote to designing deeper and wider CNN architectures for pan-sharpening. Yang \etal \cite{PanNet} design a 13-layer CNN model. They manage to build a bridge between the residual learning \cite{ResNet} and the pan-sharpening, in which the network learns to predict high-frequency details that are actually residual images between the LR MS and HR MS. Yuan \etal \cite{MSDCNN} combine a deep CNN with a shallow one to form a multi-scale solution. Liu \etal \cite{TFNet} design a two-stream network to extract enhanced modality-specific features from the PAN and MS input, respectively. Following this work, PSGAN \cite{PSGAN} further improve its performance with the adversarial learning \cite{GAN}. 

\subsection{Vision Transformers}
Transformer was first proposed by Vaswani \etal \cite{transformer} for machine translation task and then become a mainstream architecture in most of NLP tasks. Inspired by this, researchers have tried to design vision Transformers and obtained significant improvements over CNN architectures in various computer vision tasks, including but not limited to image classification \cite{ViT} and image generation \cite{ImageTransformer}. Dosovitskiy \etal \cite{ViT} design a Transformer that applied directly to sequences of image patches and achieve impressive performance on image classification. Parmar \etal \cite{ImageTransformer} extend Transformer to a sequence modeling formulation and show its effectivity in modeling textual sequence. More recently, Liu \etal \cite{Swin} propose a Transformer based vision backbone, which gains a great breakthrough by surpassing the previous state-of-the-art backbone networks by a large margin on a broad range of vision tasks. They highlight the strong potential of Transformer-like architectures for unified modeling between vision and language. Motivated by this, our design of architecture is devoted to capturing correlated and complementary information between the PAN and MS modalities.


\section{Method}
\label{sec:method}

In this section, we introduce PanFormer, a transformer based model for the pan-sharpening task. Fig.~\ref{fig::architecture} illustrates the architecture of PanFormer, which consists of 3 modules: the modality-specific encoder, the cross-modality fusion module, and the image restoration module. 

\subsection{Modality-specific Encoder}
We construct a dual-path encoder for modality-specific feature extraction, where each path processes one of the modalities. Images are first split into non-overlapped patches before feeding into the network and then projected into a hidden dimension (denoted as $C$) through a linear embedding. For the PAN image, the patch size is set as $2 \times 2$. It is transformed to a tensor of size $2H \times 2W \times C$. Considering the low resolution of the MS image, we keep its spatial resolution and only do the linear embedding so that its shape is $H \times W \times C$. Latter, each patch is treated as a token to formulate as a sequence and is processed by a series of self-attention blocks. 

\textbf{Self-attention block (SAB)} Since PAN and MS are different modalities, it is essential to extract features from them using distinct encoders. Each encoder is built with a stack of self-attention blocks to produce modality-specific intermediate features of the input. Each SAB is consist of two LayerNorm layers, one self-attention layer, and two successive MLP layers. The detailed architecture is shown in Fig.~\ref{fig::architecture}(b). The self-attention mechanism is defined as:

\begin{equation}
\label{eq:self-attention}
	\text{SA}(F) = \text{Attn}(\text{K}(F), \text{V}(F), \text{Q}(F))
\end{equation}
where $\text{SA}(\cdot)$ stands for the self-attention, $F$ is the feature vector from the previous LayerNorm, $\text{K}(\cdot), \text{V}(\cdot), \text{Q}(\cdot)$ are the linear projections for generating \textit{key}, \textit{value}, and \textit{query} (aka $K, V, Q$) vectors. The attention function $\text{Attn}(\cdot)$ has the following form:
\begin{equation}
\label{eq:attention}
	\text{Attn}(K, V, Q) = \text{SoftMax}(\frac{Q \cdot K^T}{\sqrt{C}} \cdot V)
\end{equation}

For easy training, a residual connection is applied in each LayerNorm-X module, as shown in the Fig.~\ref{fig::architecture}(b). GELU activations are applied after the 2-layer MLP.

Each modality is processed by $4$ SABs to extract modality-specific features. For the PAN path, we insert an additional Patch Merging Layer in the middle to merge patches to reach the same size as the MS path ($H \times W \times C$). Motivated by Swin Transformer \cite{Swin}, we build SAB with the window based self-attention, where the attention is computed within local windows for efficient modeling. 

\subsection{Cross-modality fusion module}
\label{fusion module}

PAN and MS images are highly correlated however contain complementary information. Thus it is important to model cross-modality relations between them. To achieve this, we develop a cross-attention block and fuse the two modalities progressively. 

\textbf{Cross-attention block (CAB)} Given two features $F_a$ and $F_b$, their relations can be modeled using attention mechanism, defined as follows, 
\begin{equation}
\label{eq:cross-attention}
	\text{CA}(F_a, F_b) = \text{Attn}(\text{K}(F_a), \text{V}(F_a), \text{Q}(F_b))
\end{equation}
where $\text{CA}(\cdot)$ is an attention function for computing relations between $F_a$ and $F_b$. We use the same attention function to Eq.~(\ref{eq:self-attention}) to compute $\text{CA}(\cdot)$.  

Specifically, there are two distinct ways to formulate the cross-attention: $K$ and $V$ are derived from the PAN (or MS) image, while $Q$ is derived from the MS (or PAN) image. We call them PAN-X-MS and MS-X-PAN attentions for convenience. The detailed influence of these two designs will be discussed in Section~\ref{sec:ablation}. The fusion module consists of $6$ CABs, as shown in Fig.~\ref{fig::architecture}(a). Here, we also employ the shifted window strategy from \cite{Swin} to achieve an effective performance. 

The outputs of different attentions are concatenated to formulate the fused representation which contains information from different modalities and are fed into the next module.

\subsection{Image restoration module}
Finally, the head aims to produce the final pan-sharpened image based on the fused representation from the cross-modality fusion module. To achieve this goal, we design a simple yet efficient restore module. There are 4 convolutional layers, each is with filter size of $3 \times 3$ and stride 1. The first two layers extract $4C$-channel feature maps, the third layer extract $C$-channel feature maps and the last one generate the final $4$-channel fusion results. Specially, the first two layers are followed by one pixel-shuffle layer, which is used to upscale the inter-mediate outputs ($\times 2$). We apply ReLU activations before each convolution layer except the first one.

\subsection{Training Details}
We use L1 loss to train our network:
\begin{equation}
	\mathcal{L} = \sum_{n=1}^{N} \| I_{ps} - I_{gt} \|_1
\end{equation}
where N is the number of training samples in a mini batch, $I_{ps}$ stands for the pan-sharpened image generated by our model, and $I_{gt}$ is the corresponding ground truth.

We implement our proposed method in PyTorch \cite{PyTorch} and train it on a single NVIDIA Titan 2080Ti GPU. Adam optimizer \cite{Adam} is used to train the model for $200K$ iterations with fixed hyper-parameters $\beta_1 = 0.9$ and $\beta_2 = 0.999$. The batch-size is set as 4 and the initial learning-rate is set as $ 1 \times 10^{-4}$. The learning-rate is multiplied by $0.99$ for every $10K$ iterations for an efficient convergence. The hidden channel $C$ is set as $64$. In SABs and CABs, the window size is set as $4$ and each attention consists of $8$ heads. The count of trainable parameters of our model is about $1.53M$. In addition, we keep all remote sensing images with their raw bit depth in both training and testing stages to be the same as the condition in the real world.  

\section{Experiments}
\label{sec:exp}

\subsection{Datasets and metrics}
We conduct extensive experiments on the remote sensing images acquired by GaoFen-2 and WorldView-3 satellites. Take GaoFen-2 dataset as an example, we choose one image as the test image and crop it orderly into $286$ sub-images with a small overlapping region between the neighboring ones. The training set is constructed by $24,000$ patches randomly cropped from other images. We repeat the same process to generate the WorldView-3 dataset, which is comprised of $10,000$ samples for training and $99$ samples for testing. Note that there is no overlap between the training and testing sets. To generate the ground truth images, we follow Wald's protocol \cite{wald}. Both PAN and MS images are down-sampled to a lower scale so that the original MS images serve as the ground truth images. Specifically, we blur the original image pairs using a Gaussian filter and down-sample them with a scaling factor of 4 to form the testing and training patches. Finally, we train the model with PAN images in size of $256 \times 256 \times 1$, LR MS images in size of $64 \times 64 \times 4$ and the ground truth HR MS images in size of $256 \times 256 \times 4$. In the testing stage, the sizes of PAN, LR MS, and HR MS images become $400 \times 400 \times 1$, $100 \times 100 \times 4$, and $400 \times 400 \times 4$. 

To evaluate our model, we select $4$ popular metrics including \textit{peak signal-to noise ratio} (PSNR), \textit{structural similarity} (SSIM) \cite{SSIM}, \textit{relative dimensionless global error in synthesis} (ERGAS) \cite{ERGAS}, and \textit{spatial correlation coefficient} (SCC) \cite{sCC}. The result is better if its PSNR, SSIM and SCC are higher, and ERGAS is lower.

\begin{figure}[t]
  \centering
  \begin{minipage}[b]{.49\linewidth}
    \includegraphics[width=\linewidth]{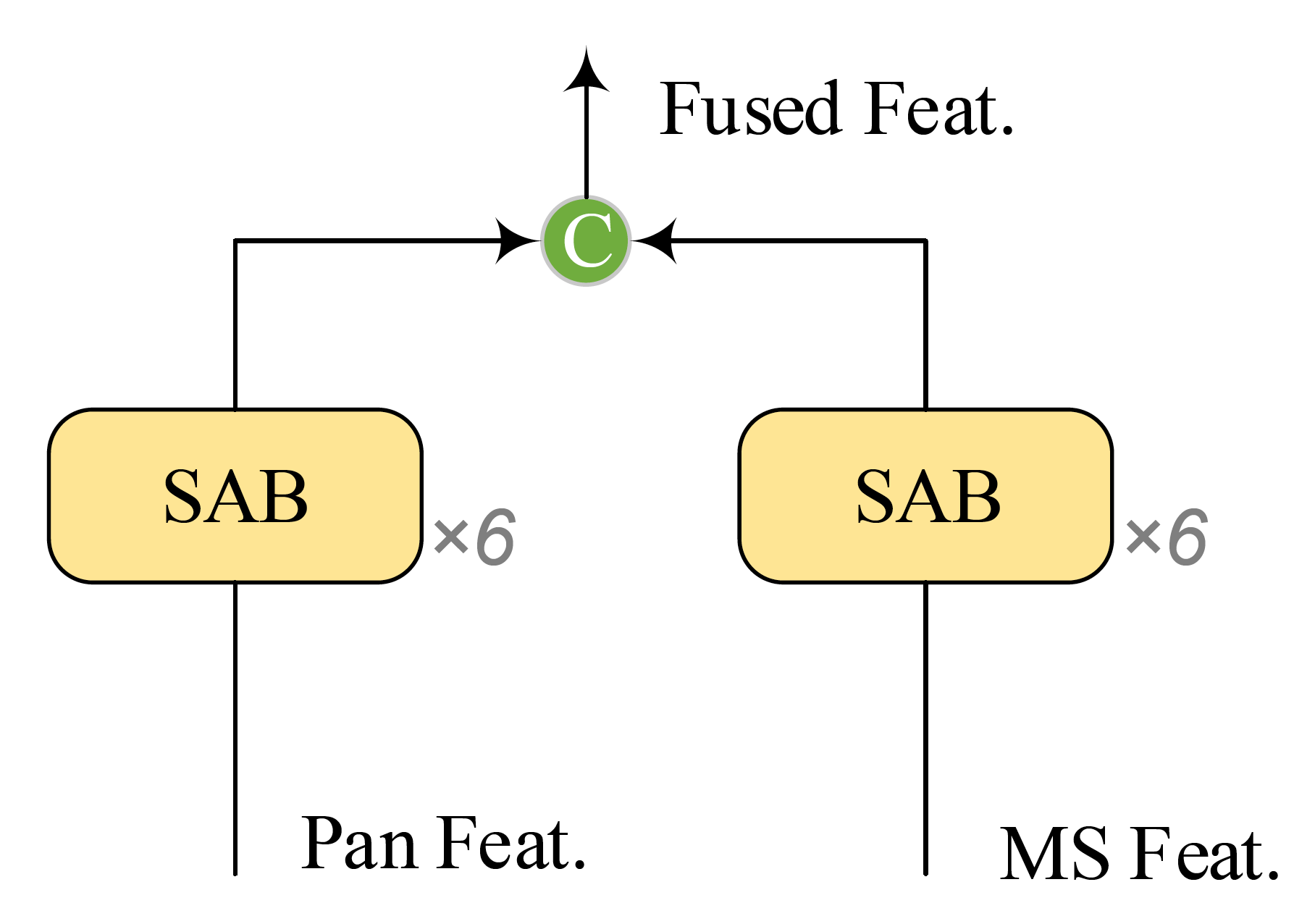}
    \centerline{(a) Concat}
  \end{minipage}
  \begin{minipage}[b]{.49\linewidth}
    \includegraphics[width=\linewidth]{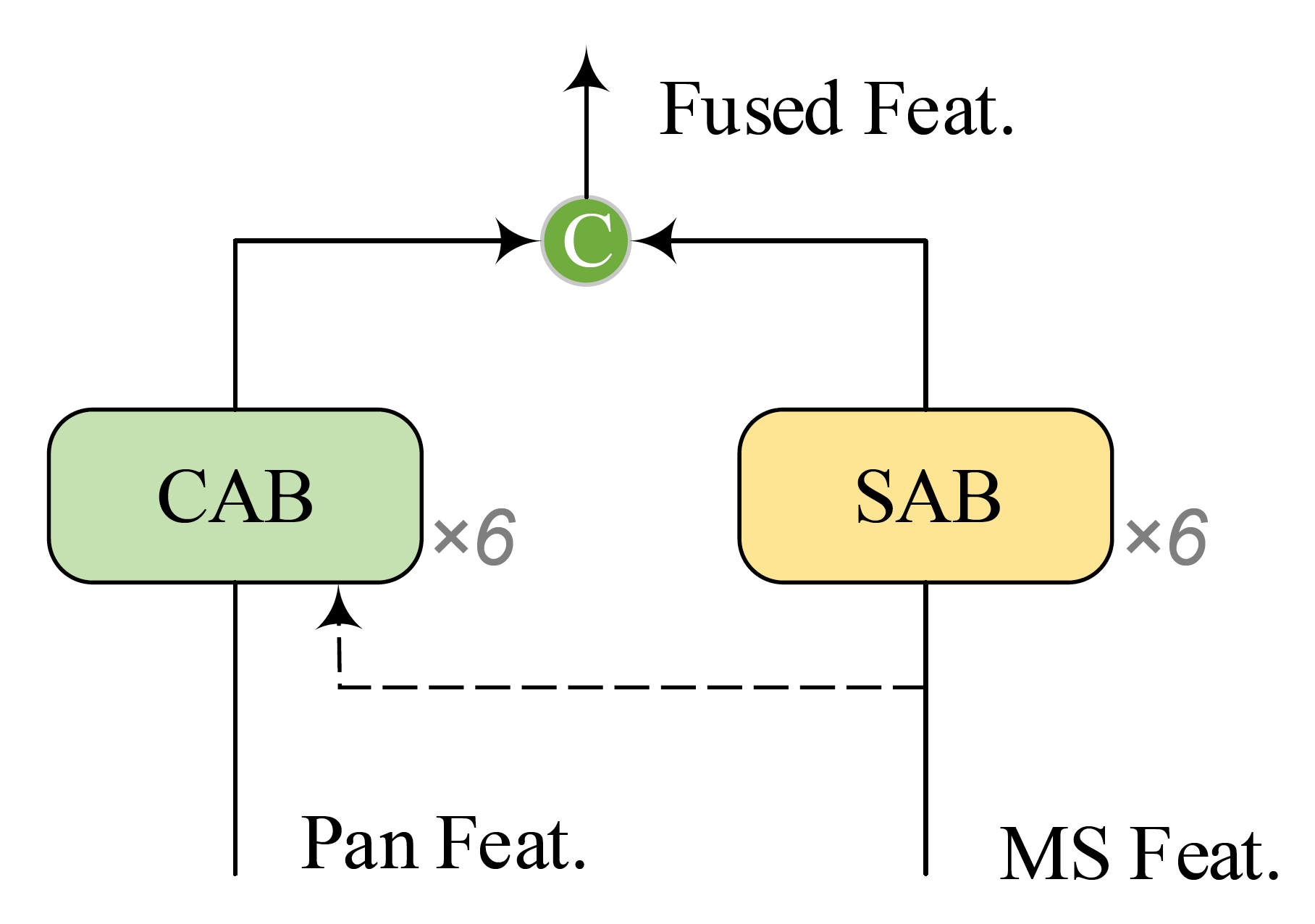}
    \centerline{(b) PAN-X-MS}
  \end{minipage}
  
  \begin{minipage}[b]{.49\linewidth}
    \includegraphics[width=\linewidth]{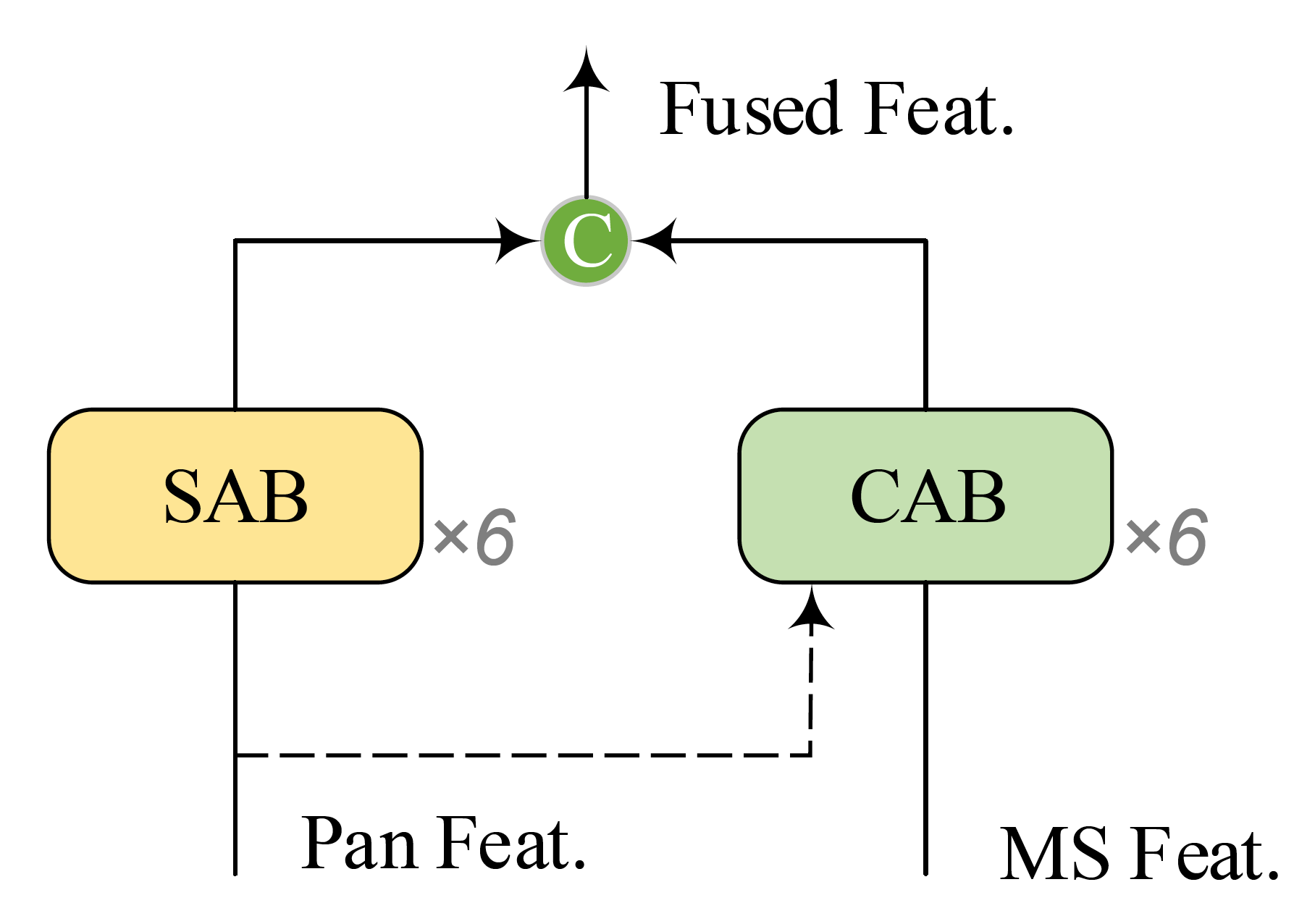}
    \centerline{(c) MS-X-PAN}
  \end{minipage}
  \begin{minipage}[b]{.49\linewidth}
    \includegraphics[width=\linewidth]{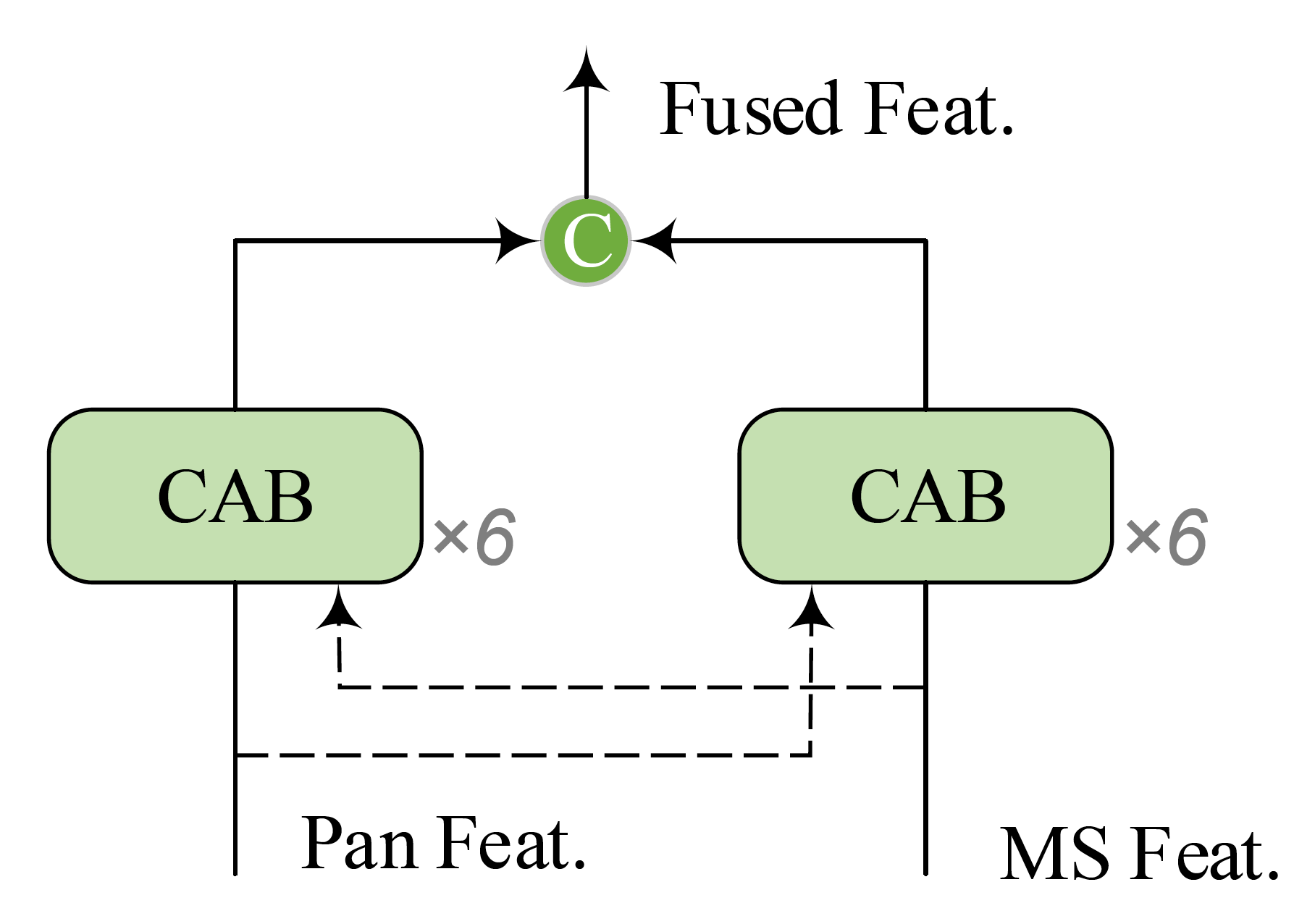}
    \centerline{(d) PanFormer}
  \end{minipage}
  \caption{Four different ways to implement the feature fusion module.}
  \label{fig::ablation}
  \vspace{-10pt}
\end{figure}

\begin{table}[t]
\centering
  \caption{Ablation study on different fusion strategies. Detailed architectures for completing fusion are depicted in Fig.~\ref{fig::ablation}. The best results are highlighted in \textbf{bold}.}
  \vspace{0pt}
  \label{tab::ablation}
  \begin{tabular}{l|c|c|c|c}
    \toprule
    Model & PSNR $\uparrow$ & SSIM $\uparrow$ & ERGAS $\downarrow$ & SCC $\uparrow$ \\
    \midrule
    Concat & 41.0963 & 0.9737 & 1.2364 & 0.9734 \\
    PAN-X-MS & 41.3788 & \textbf{0.9752} & 1.1824 & 0.9736 \\
    MS-X-PAN & 40.8486 & 0.9738 & 1.2532 & \textbf{0.9738} \\
    PanFormer & \textbf{41.4281} & \textbf{0.9752} & \textbf{1.1748} & 0.9735 \\
	\bottomrule
\end{tabular}
\vspace{-10pt}
\end{table}

\begin{table*}[t]
\centering
  \caption{The quantitative results on different satellites. The best results are highlighted in \textbf{bold} and the second is \underline{underlined}. }
  \vspace{5pt}
  \label{tab::results}
  \begin{tabular}{l|cccc|cccc}
    \toprule
     & \multicolumn{4}{c|}{GaoFen-2} & \multicolumn{4}{c}{WorldView-3} \\
     & PSNR $\uparrow$ & SSIM $\uparrow$ & ERGAS $\downarrow$ & SCC $\uparrow$ & PSNR $\uparrow$ & SSIM $\uparrow$ & ERGAS $\downarrow$ & SCC $\uparrow$ \\
    \midrule
    BDSD~\cite{BDSD}	& 27.4540 &	0.6969 & 6.2093	& 0.9458 & 31.0816 & 0.8493 & 5.5417 & 0.9276 \\
	GS~\cite{GS}		& 31.6249 & 0.8383 & 3.7190	& 0.9470 & 30.0437 & 0.8309 & 6.2308 & 0.9232 \\
	PNN~\cite{PNN}		& 39.4107 & 0.9626 & 1.4982	& 0.9657 & \underline{33.7584} & 0.9250 & 4.1083 & 0.9405 \\
	MSDCNN~\cite{MSDCNN}	& 39.5169 & 0.9629 & 1.4904 & 0.9599 & 32.4680 & 0.9225 & 4.2856 & 0.9414 \\
	DRPNN~\cite{DRPNN}	& 40.7058 & 0.9708 & 1.2838	& 0.9694 & 33.1806 & 0.9301 & 4.5375 & 0.9165 \\
	PanNet~\cite{PanNet}	& 40.6766 & 0.9712 & 1.2923	& 0.9728 & 33.6837 & 0.9180 & 4.3263 & 0.9494 \\
	PSGAN~\cite{PSGAN}	& \underline{41.1466} & \underline{0.9745} & \underline{1.2092}	& \textbf{0.9742} & 33.1308 & 0.9311 & 4.1070 & \textbf{0.9628} \\
	GPPNN~\cite{GPPNN}	& 39.5548 & 0.9650 & 1.4780 & 0.9623 & 33.6900 & \textbf{0.9451} & 3.8711 & \underline{0.9559} \\
	PanFormer & \textbf{41.4281} & \textbf{0.9752} & \textbf{1.1748} & \underline{0.9735} & \textbf{34.4175} & \underline{0.9445} & \textbf{3.6746} & 0.9495 \\
	\bottomrule
\end{tabular}
\vspace{-2pt}
\end{table*}

\begin{figure*}[t]
  \centering
  \begin{minipage}{.15\linewidth}
    \includegraphics[width=\linewidth]{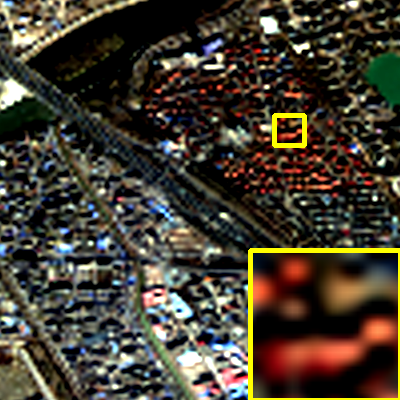}
    \centerline{(a) LR MS}
  \end{minipage}
  \begin{minipage}{.15\linewidth}
    \includegraphics[width=\linewidth]{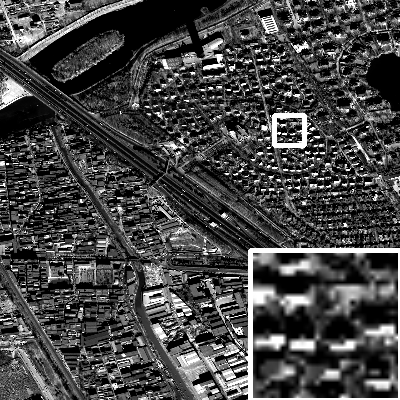}
    \centerline{(b) PAN}
  \end{minipage}
  \begin{minipage}{.15\linewidth}
    \includegraphics[width=\linewidth]{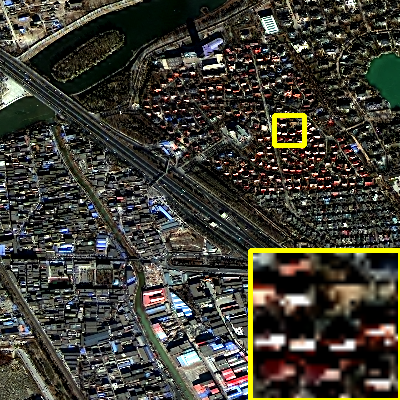}
    \centerline{(c) BDSD \cite{BDSD}}
  \end{minipage}
  \begin{minipage}{.15\linewidth}
    \includegraphics[width=\linewidth]{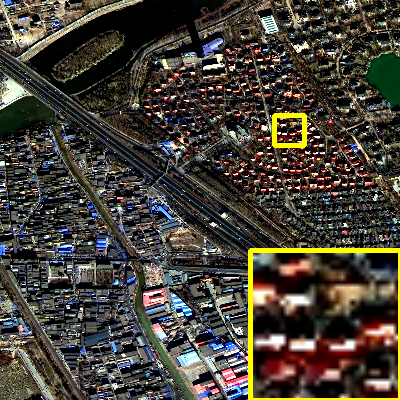}
    \centerline{(d) GS \cite{GS}}
  \end{minipage}
  \begin{minipage}{.15\linewidth}
    \includegraphics[width=\linewidth]{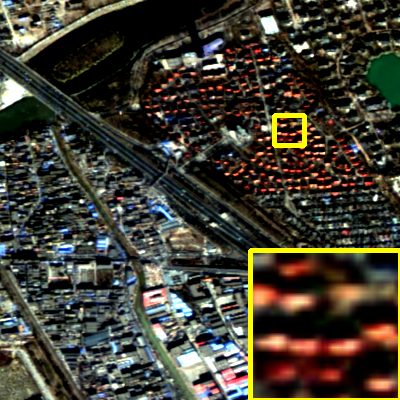}
    \centerline{(e) PNN \cite{PNN}}
  \end{minipage}
  \begin{minipage}{.15\linewidth}
    \includegraphics[width=\linewidth]{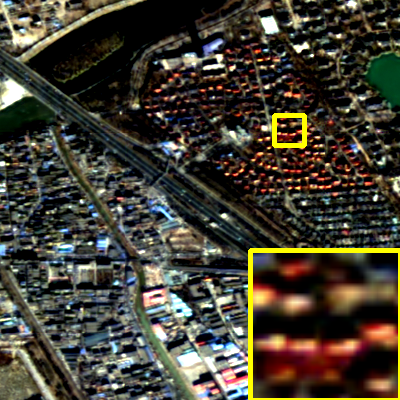}
    \centerline{(f) MSDCNN \cite{MSDCNN}}
  \end{minipage}
  \begin{minipage}{.15\linewidth}
    \includegraphics[width=\linewidth]{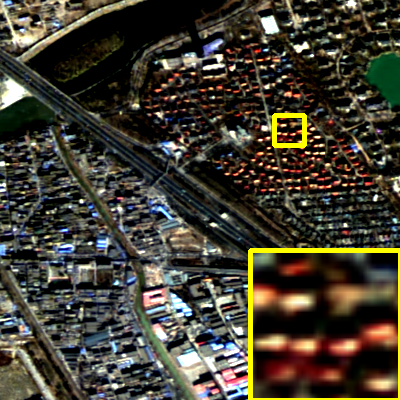}
    \centerline{(g) PanNet \cite{PanNet}}
  \end{minipage}
  \begin{minipage}{.15\linewidth}
    \includegraphics[width=\linewidth]{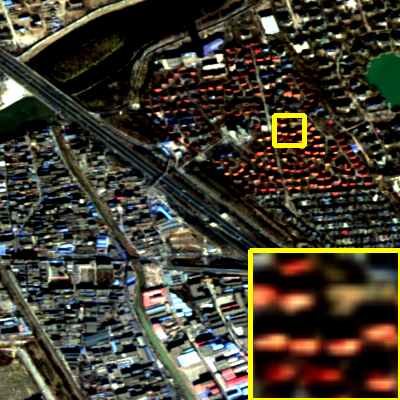}
    \centerline{(h) PSGAN \cite{PSGAN}}
  \end{minipage}
  \begin{minipage}{.15\linewidth}
    \includegraphics[width=\linewidth]{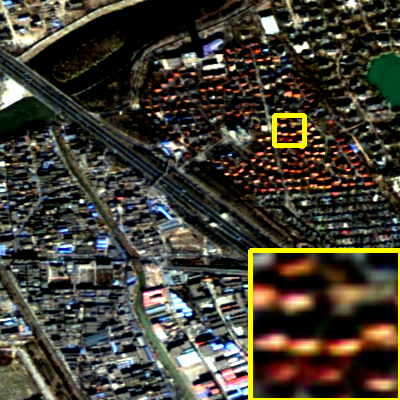}
    \centerline{(i) DRPNN \cite{DRPNN}}
  \end{minipage}
  \begin{minipage}{.15\linewidth}
    \includegraphics[width=\linewidth]{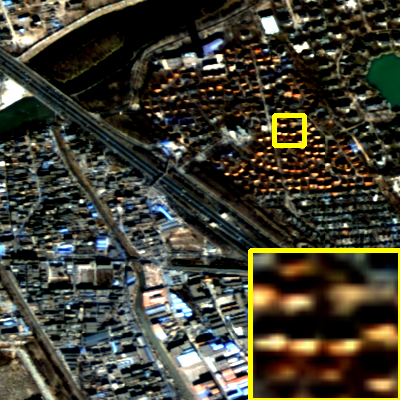}
    \centerline{(j) GPPNN \cite{GPPNN} }
  \end{minipage}
  \begin{minipage}{.15\linewidth}
    \includegraphics[width=\linewidth]{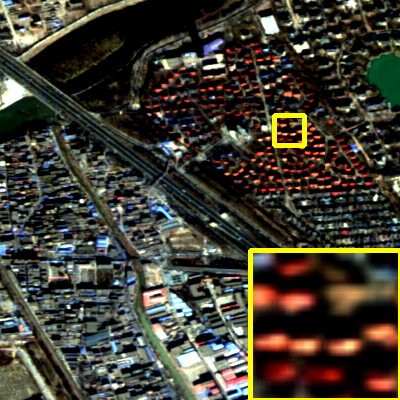}
    \centerline{(k) PanFormer}
  \end{minipage}
  \begin{minipage}{.15\linewidth}
    \includegraphics[width=\linewidth]{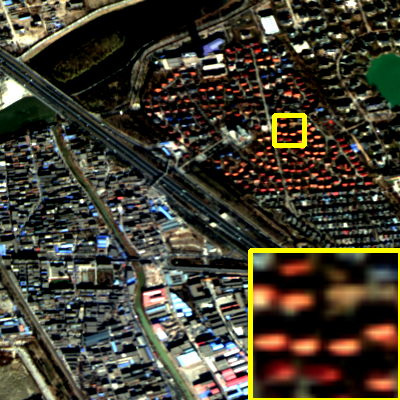}
    \centerline{(l) GT}
  \end{minipage}
  \vspace{-5pt}
  \caption{Examples of visual results on GaoFen-2 satellite. Visualized in RGB.}
  \label{fig::GF-2}
\vspace{-5pt}
\end{figure*}

\subsection{Effects of the fusion module} 
\label{sec:ablation}
Fusion module plays a critical role in pan-sharpening. In this section, we test different implementations of fusion module in PanFormer. The detailed architectures of them are depicted in Fig.~\ref{fig::ablation}. There are $4$ ways to fuse a pair of features: 
\begin{itemize}
	\item \textbf{Concat} simply concatenate the features from different modalities. 
	\begin{equation}
		F_{fuse} = [\text{SA}(F_{pan}); \text{SA}(F_{ms})]
	\end{equation}
	where $F_{pan}, F_{ms}$ are features extracted by the encoders from PAN and MS images, respectively. $[\cdot; \cdot]$ means the concatenate. $F_{fuse}$ is the fused feature.
	\item \textbf{PAN-X-MS} use CABs to fuse the features, where PAN features is used to generate $K$ and $V$, and MS features is used to generate $Q$. 
	\begin{equation}
		F_{fuse} = [\text{CA}(F_{pan}, F_{ms}); \text{SA}(F_{ms})]
	\end{equation}
	\item \textbf{MS-X-PAN} is similar to PAN-XS-MS. The difference is that MS features is used to generate $K$ and $V$, and PAN features is used to generate $Q$.
	\begin{equation}
		F_{fuse} = [\text{SA}(F_{pan}); \text{CA}(F_{ms}, F_{pan})]
	\end{equation}
	\item \textbf{PanFormer} use both PAN-X-MS and MS-X-PAN attentions and concatenate their results as the fused features.
	\begin{equation}
		F_{fuse} = [\text{CA}(F_{pan}, F_{ms}); \text{CA}(F_{ms}, F_{pan})]
	\end{equation}
\end{itemize} 
To balance the count of parameters in these variants, we add the same number of SABs for the path without CABs.

The results are displayed in Table~\ref{tab::ablation}. We can see that PAN-X-MS can improve the performance of the simple baseline Concat, while the improvement of MS-X-PAN is marginal and even worse in some metrics. However, when we combine these two cross-attention strategies, it achieves the best results. It is demonstrated that the exchange of information between different modalities during the cross-attention module is helpful to generate better fused representations. We conclude that the cross-modality fusion module does improve the quality of the final pan-sharpened images.

\subsection{Comparisons with SOTA methods}
Evaluations on different satellites are carried out to evaluate the performance of PanFormer. We select $6$ SOTA deep learning based methods including: PNN \cite{PNN}, MSDCNN \cite{MSDCNN}, PanNet \cite{PanNet}, PSGAN \cite{PSGAN}, DRPNN \cite{DRPNN}, and GPPNN \cite{GPPNN}, and $2$ traditional methods including BDSD \cite{BDSD}, and GS \cite{GS} for comparison.

The quantitative results are reported in Table~\ref{tab::results}. It is noted that our proposed model is the only one that based on Transformer. From the results, we can see that PanFormer achieves the best PSNR, SSIM and ERGAS values, and the second best SCC value on GaoFen-2 dataset, which shows a great competitive ability of PanFormer to those CNN based models. When evaluated on WorldView-3 dataset, the performances of all deep learning based methods drop due to that images captured by WorldView-3 contain more MS bands (8 bands) and there are fewer data available in our training set. However, PanFormer still obtains the best SSIM and ERGAS values, and the second best SCC value on WorldView-3 dataset.  

\begin{figure*}[t]
\centering
  \begin{minipage}[b]{.135\linewidth}
    \includegraphics[width=\linewidth]{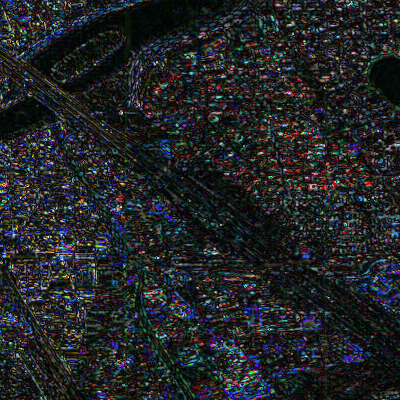}
    \centerline{(b) PNN \cite{PNN}}
  \end{minipage}
  \begin{minipage}[b]{.135\linewidth}
    \includegraphics[width=\linewidth]{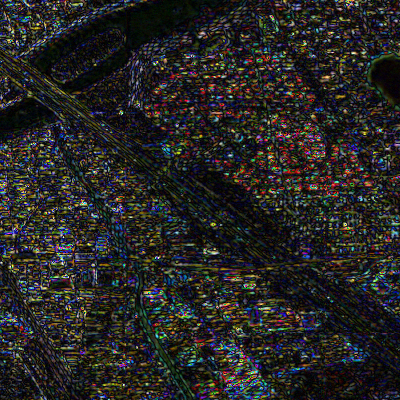}
    \centerline{(c) MSDCNN \cite{MSDCNN}}
  \end{minipage}
  \begin{minipage}[b]{.135\linewidth}
    \includegraphics[width=\linewidth]{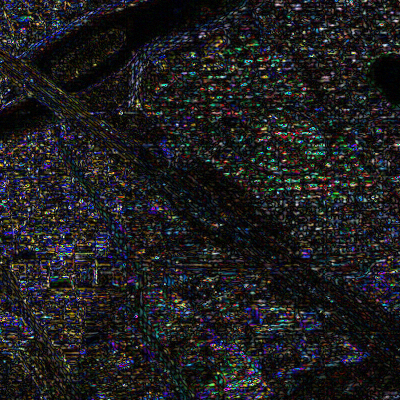}
    \centerline{(d) PanNet \cite{PanNet}}
  \end{minipage}
  \begin{minipage}[b]{.135\linewidth}
    \includegraphics[width=\linewidth]{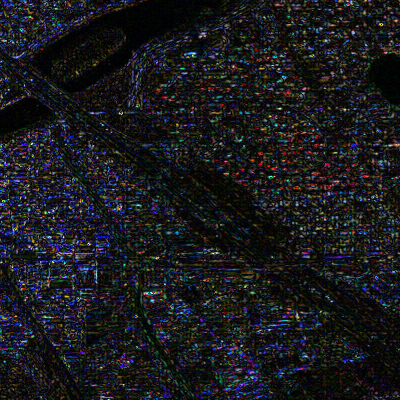}
    \centerline{(e) PSGAN \cite{PSGAN}}
  \end{minipage}
  \begin{minipage}[b]{.135\linewidth}
    \includegraphics[width=\linewidth]{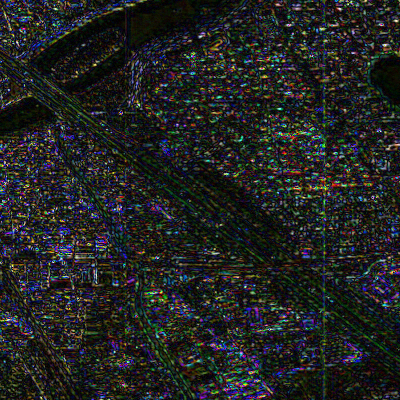}
    \centerline{(f) DRPNN \cite{DRPNN}}
  \end{minipage}
  \begin{minipage}[b]{.135\linewidth}
    \includegraphics[width=\linewidth]{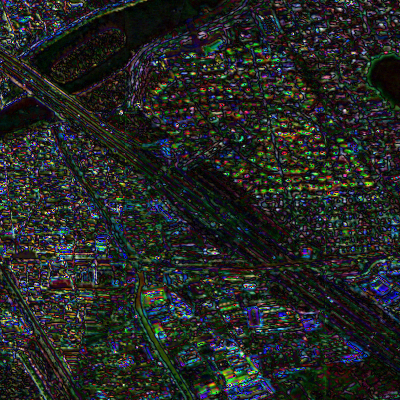}
    \centerline{(g) GPPNN \cite{GPPNN} }
  \end{minipage}
  \begin{minipage}[b]{.135\linewidth}
    \includegraphics[width=\linewidth]{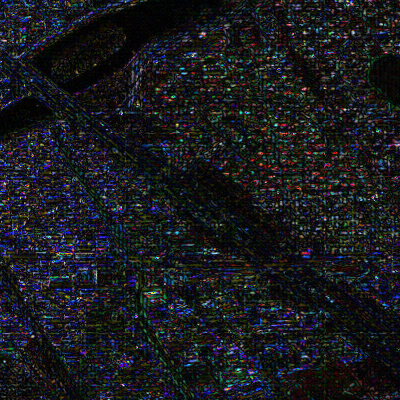}
    \centerline{(h) PanFormer}
  \end{minipage}
  \vspace{-10pt}
  \caption{The residuals between the HR MS image reconstructions and the ground truth from Fig.~\ref{fig::GF-2} .}
  \label{fig::GF-2-Res}
\vspace{-10pt}
\end{figure*}

We visualize some examples of different satellites to evaluate the performance of models in reality applications. Fig.~\ref{fig::GF-2} display some representative results on GaoFen-2 dataset. We also provide the corresponding residuals in Fig.~\ref{fig::GF-2-Res}. On Gaofen-2 dataset, the traditional methods BDSD (Fig.~\ref{fig::GF-2}(c)) and GS (Fig.~\ref{fig::GF-2}(d)) generate results with great noises. The deep learning based methods PanNet (Fig.~\ref{fig::GF-2}(g)), DRPPN (Fig.~\ref{fig::GF-2}(i)) and GPPNN (Fig.~\ref{fig::GF-2}(j)) suffer from color distortions. We can still find PNN (Fig.~\ref{fig::GF-2-Res}(b)), MSDCNN (Fig.~\ref{fig::GF-2-Res}(c)) produce more discrepancies from the residuals in Fig~\ref{fig::GF-2-Res}. Our method ((Fig.~\ref{fig::GF-2}(q)) produces the closet result to the GT (Fig.~\ref{fig::GF-2}(r)) with smallest spatial or spectral distortions.

We report the inference time and the size of deep learning based models in Table~\ref{tab::efficiency}. The average time cost per image in the WorldView-3 testing set is given and only the parameters in the generator are count for those GAN based model. Table~\ref{tab::efficiency} shows that our method share a similar parameter count to DRPNN \cite{DRPNN} and PSGAN \cite{PSGAN}. The time cost of our method is slower than those CNN based methods because the attention calculation is time-consuming compared to convolutions.

\begin{table}[t]
\centering
  \caption{Inference time and number of parameters of deep models. Note that the pan-sharpened images are with size of $400 \times 400 \times 4$. We give an average time of them. As for GAN based models, we only count the parameters in the generator.}
  \vspace{5pt}
  \label{tab::efficiency}
  \begin{tabular}{r|c|c}
    \toprule
    Model					& $\sharp$Params & Time(s)\\
    \midrule
    PNN \cite{PNN} 			& 0.080M		& 0.0012 \\
    MSDCNN \cite{MSDCNN} 	& 0.262M		& 0.0035 \\
    DRPNN \cite{DRPNN} 		& 1.639M		& 0.0031 \\
    PanNet \cite{PanNet} 	& 0.078M		& 0.0032 \\
    PSGAN \cite{PSGAN} 		& 1.654M		& 0.0045 \\
    GPPNN \cite{GPPNN} 		& 0.120M		& 0.0211 \\
    PanFormer 				& 1.530M		& 0.0468 \\
    \bottomrule
\end{tabular}
\end{table}

\section{Conclusion}
\label{sec:con}
In this paper, we propose a novel pan-sharpening model based on Transformer, termed as PanFormer. To the best of our knowledge, it is the first attempt to introduce Transformer to the pan-sharpening problem. The self-attention blocks extract the modality-specific features from PAN and MS images, respectively. The cross-attention blocks help us to fuse enhanced features. We explore different settings for the attention mechanism and conduct ablation studies to verify the rationality of our module. Experiments on GaoFen-2 and WorldView-3 datasets demonstrate that PanFormer can produce impressive fine-grained pan-sharpened results which show its competitive ability to the existed CNN based methods. 

\small
\bibliographystyle{IEEEbib}
\bibliography{icme2022template}

\end{document}